\documentclass[conference]{IEEEtran}
\IEEEoverridecommandlockouts
\usepackage{cite}
\usepackage{amsmath,amssymb,amsfonts}
\usepackage{algorithmic}
\usepackage{graphicx}
\usepackage{subcaption}
\usepackage{textcomp}
\usepackage{xcolor}
\usepackage{makecell}
\usepackage{multirow}
\usepackage{latexsym}
\usepackage{booktabs}
\usepackage{color}
\usepackage{hyperref}
\usepackage{ragged2e} 
\usepackage{mathtools}
\usepackage{comment}
\usepackage{threeparttable}
\def\BibTeX{{\rm B\kern-.05em{\sc i\kern-.025em b}\kern-.08em
    T\kern-.1667em\lower.7ex\hbox{E}\kern-.125emX}}

\begin{document}

% \title{Acting Two Using Only One: A Novel Framework for Extending Single-Arm Agent Actions to Dual Arms}
\title{One-to-Two Acting: A Novel Framework for Single-arm Agent Action Expansion to Dual Arms}

% \author{Anonymous ICME submission}
\author{
Youbin Yao$^{1,*}$, Nieqin Cao$^{2,*}$, Mingyan Li$^{1}$, Yan Ding$^{3}$, Fuqiang Gu$^{1}$, and Chao Chen$^{1,\dagger}$\\
$^{1}$Chongqing University, Chongqing, China $^{2}$Xi'an Jiaotong-Liverpool University, Suzhou, China
$^{3}$Lumos Roboticss\\
% \texttt{yaoyoubin@stu.cqu.edu.cn, Nieqing.Cao@xjtlu.edu.cn, \\limy2021@cqu.edu.cn, yding25@binghamton.edu, \{gufq, cschaochen\}@cqu.edu.cn}\\
\texttt{cschaochen@cqu.edu.cn}\\
\thanks{$^{*}$Equal contribution \qquad $^{\dagger}$Corresponding author}
\vspace{-5mm}
}
\maketitle

\begin{abstract}
Dual-arm manipulation can improve throughput via parallel execution, but collecting bimanual demonstrations for training is costly and difficult. We present ExS2D (\underline{Ex}tending \underline{S}ingle-Arm Agent Actions \underline{to} \underline{D}ual Arms), a hierarchical 
%dual-arm 
action expansion framework that enables dual-arm manipulation from single-arm supervision. ExS2D first generates structured subtasks according to textual instructions while explicitly capturing their temporal precedence. Then, ExS2D grounds each subtask into executable actions via subtask-guided action mapping in observation. Finally, precedence-aware action allocation and synchronized planning are then performed by an Multimodal Large Language Model‌ (MLLM) driven coordinator so as to select collision-free dual-arm executions. 
Simulation experiments demonstrate that ExS2D reduces the average execution steps by $54.4\%$, while maintaining a comparable success rate to a single-arm baseline. Real-robot experiments on four tasks further demonstrate the ExS2D's  reliability in terms of dual-arm execution under few-shot single-arm samples, while using zero bimanual demonstrations.
\end{abstract}


\section{Introduction}
Dual-arm manipulators can substantially improve throughput and versatility for tabletop rearrangement~\cite{JUNGBLUTH2022156} by enabling parallel execution and bimanual coordination~\cite{jiang2023vimageneralrobotmanipulation,wu2024fastumiscalablehardwareindependentuniversal}. 
However, scaling dual-arm policies remains difficult in practice because collecting high-quality bimanual datasets is costly and many approaches require additional domain engineering or strong assumptions to coordinate two arms reliably~\cite{10900471, zhao2023learningfinegrainedbimanualmanipulation,chu2025llmmapbimanualrobottask}. Worse still, existing dual-arm manipulation policies~\cite{zhao2023learningfinegrainedbimanualmanipulation, 10323148} rely heavily on predefined rigid strategies, lacking flexibility to adapt to real-time changes~\cite{wang2025exploring} (e.g., arm malfunctions or task irregularities).
%and underscoring the need of a more dynamic and flexible policy
In contrast, single-arm agents  have matured with abundant single-arm samples supervision, offering low-level actions primitives that are readily reusable.

A natural question is whether we can upgrade a capable single-arm agent to dual-arm execution without relying on bimanual datasets. 
Many existing methods trained with abundant single-arm data, such as CLIPort~\cite{pmlr-v164-shridhar22a}, while the current single-arm agents are fundamentally designed for sequential single-action generation~\cite{joublin2023copalcorrectiveplanningrobot,jiang2023vimageneralrobotmanipulation, su2025ova}. 
This limitation results in efficiency issues when handling rearrangement tasks that can be performed simultaneously by multiple manipulators~\cite{li2024synchronizeddualarmrearrangementcooperative,zhang2024learningdualarmobjectrearrangement}. 
The primary reason for such inefficiency is that existing single-arm agents do not account for inter-action dependency, spatial relationship, and synchronized coordination between arms.
To systematically extend such single-arm oepration to dual-arm execution, we must address three key challenges:
% (i) decomposing long-horizon instructions into structured subtasks,
% (ii) grounding each subtask into precise action primitives, and
% (iii) allocating multiple actions across two arms while respecting temporal precedence and collision constraints.
(i) \textbf{Subtask structuring:} Long-horizon instructions are underspecified, making it difficult to derive a complete and temporally consistent subtask set;
(ii) \textbf{Action grounding:} Mapping each subtask to precise action primitives requires reliable object grounding  under  ambiguity;
(iii) \textbf{Dual-arm coordination:} Allocating multiple actions across two arms must satisfy precedence constraints while avoiding collisions.

In this work, we present \textbf{ExS2D} (\underline{Ex}tending \underline{S}ingle-arm agent actions \underline{to} \underline{D}ual arms), a hierarchical framework that turns single-arm supervision into coordinated dual-arm execution. ExS2D first uses an \textbf{VL-SubGen} (\underline{V}ision-\underline{L}anguage \underline{Sub}task \underline{Gen}erator) to generate a structured subtask sequence. It then grounds each subtask into executable action primitives via \textbf{SA-Map} (\underline{S}ubtask-Guided \underline{A}ction \underline{Map}ping), which combines Vision Language Model (VLM) driven localization and object mask with CLIPort to suppress irrelevant regions and refine action affordances. Finally, a \textbf{P-DCoord} (\underline{P}recedence-aware \underline{D}ual-arm \underline{Coord}inator) reasons over the subtask set, ranks candidate actions with a lightweight motion-cost heuristic, and verifies feasibility with synchronized dual-arm motion planning  and collision checking before execution.

We evaluate ExS2D on both language-conditioned task simulation and real dual-arm experiments with Elephant manipulators.
% Results show that ExS2D reduces execution steps compared to a single-arm baseline while maintaining competitive success, and transfers to real-world tasks using only few-shot single-arm samples with zero bimanual demonstrations.
In summary, our main contributions are concluded as follows:
1) \textbf{Hierarchical dual-arm framework.} We propose a hierarchical dual-arm framework that decomposes tasks into structured subtasks and explicitly reasons about precedence dependencies for coordinated execution.
2) \textbf{Efficient and reliable execution.} \textbf{P-DCoord} performs precedence-aware, motion-cost-minimizing dual-arm pairing, together with \textbf{SA-Map}'s mask-guided feasibility checks for robust grounding.
3) \textbf{Zero bimanual data collection:} All dual-arm results in experiments are obtained using no bimanual demonstrations  for training. \textbf{ExS2D} reduces average simulation execution steps by 54.4\% with comparable success rates.

\section{related work}
\subsection{Single-Arm Manipulation}
Recent work has shown that single-arm manipulation policies can generate executable action sequences by leveraging the commonsense reasoning and contextual understanding of Large Language Mode (LLM)s~\cite{zhang2023lohoravenslonghorizonlanguageconditionedbenchmark,liu2023llmpempoweringlargelanguage,11075556,yang2024bestmanmodularmobilemanipulator}. 
High-level language planning has been combined with reinforcement learning and motion planning to bridge abstract instructions and low-level control~\cite{liu2026fly}, while multimodal prompting further improves generalization across manipulation tasks~\cite{jiang2023vimageneralrobotmanipulation}. 
These approaches demonstrate the versatility and scalability of single-arm agents.

However, single-arm policies inherently suffer from efficiency limitations in tasks requiring parallel execution, as actions are restricted to sequential operation. 
Although methods such as LLM+P~\cite{liu2023llmpempoweringlargelanguage} and CoPAL~\cite{joublin2023copalcorrectiveplanningrobot} extend single-arm planning to dual-arm settings, the resulting plans remain insufficient for simultaneous multi-object manipulation. 
Moreover, single-arm agents typically rely on pre-trained manipulation skills obtained via imitation learning or base skill priors~\cite{10685120}, and systems such as CyberDemo~\cite{Wang_2024_CVPR} further enhance dexterous control through simulated demonstrations and data augmentation. 
Given that single-arm agents already possess these transferable skills, directly extending them to dual-arm execution avoids costly dual-arm data collection.
% offers a promising path to avoid costly dual-arm data collection.

\subsection{Dual-Arm Manipulation}
Most dual-arm manipulation policies rely on imitation learning with predefined bimanual datasets, whose collection is expensive and challenging. 
Teleoperation interfaces such as 3D spacemouse~\cite{chi2023diffusion}, VR controllers~\cite{pmlr-v164-jang22a}, and haptic devices~\cite{pmlr-v205-shaw23a} enable data acquisition but suffer from high cost, latency, or limited usability. 
Leader–follower systems, 
including ALOHA~\cite{zhao2023learningfinegrainedbimanualmanipulation} and GELLO~\cite{wu2024gellogenerallowcostintuitive}, 
provide more intuitive and low-cost alternatives but are largely restricted to static setups, while Mobile ALOHA~\cite{fu2024mobilealohalearningbimanual} and exoskeleton-based systems~\cite{10323148} improve environmental coverage at the cost of increased hardware complexity and data-collection effort. 
 
Recent LLM-driven approaches, such as Two-Step~\cite{bai2024twostep} and LLM+MAP~\cite{chu2025llmmapbimanualrobottask}, generate partial-order PDDL plans from language and scene descriptions to enable parallel execution, yet lack the fine-grained control needed for precision-critical manipulation. 
RoCo~\cite{10610855} reduces planning latency via learned motion-cost estimation but remains limited by coarse heuristics in tightly constrained workspaces. 
% Overall, these methods depend on explicit environment representations or domain-specific descriptions and struggle to achieve end-to-end, perception-driven dual-arm manipulation, motivating the unified design of ExS2D.
These methods rely on explicit environment representations or domain-specific descriptions, limiting end-to-end, perception-driven dual-arm manipulation and motivating ExS2D.

\begin{figure*}
\centering
\includegraphics[width=0.97\linewidth]{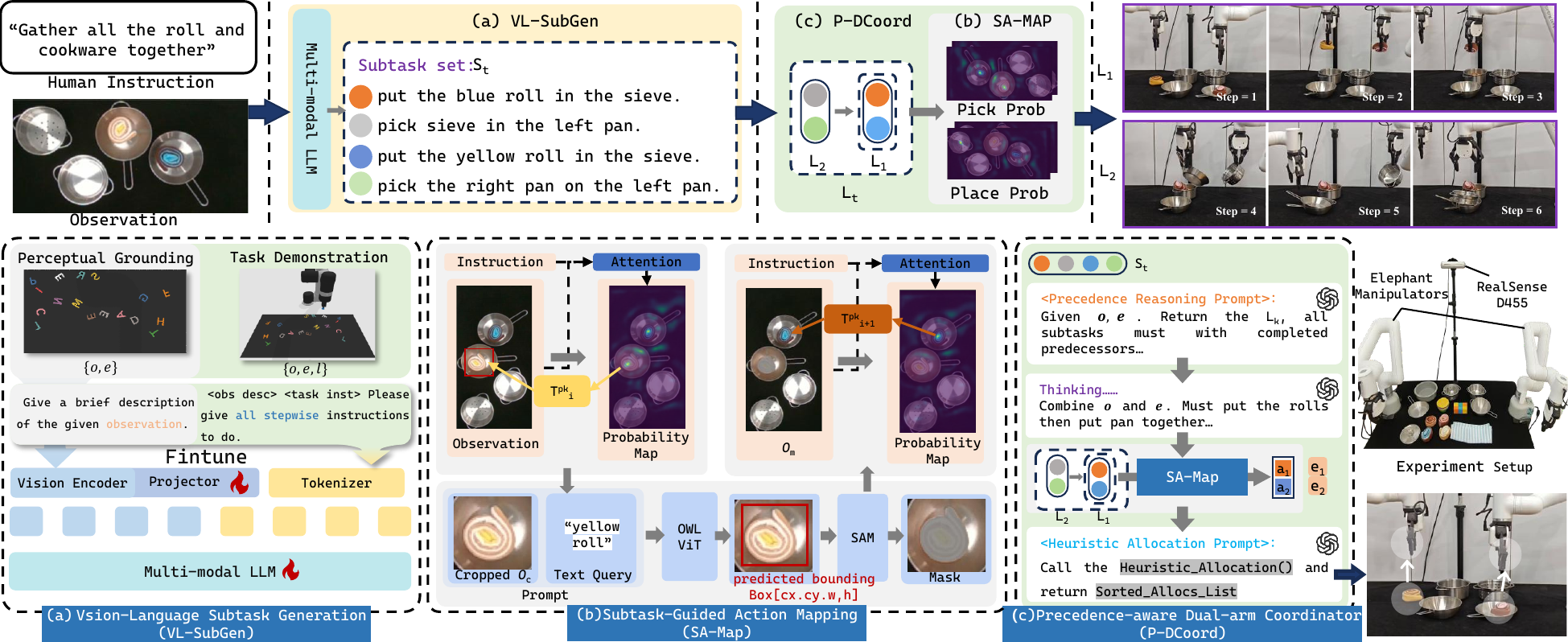}
\caption{\textbf{Framework overview.} ExS2D executes hierarchical dual-arm manipulation in three steps: (i) \textbf{VL-SubGen} generates structured $\mathcal{S}_t$ from observations and instructions (Section~\ref{sec:4.1}); (ii) \textbf{SA-Map} grounds each subtask into action primitives using OWL-ViT and SAM to produce action maps (Section~\ref{sec:4.2}); (iii) \textbf{P-DCoord} enforces precedence and motion constraints to form $\{L_k\}$ and select feasible action pairs (Section~\ref{sec:4.3}). \textbf{Top:} System pipeline. \textbf{Bottom:} Core modules and experiment  setup.}

\label{fig:1}
\vspace{-5mm}
\end{figure*}

\section{PROBLEM STATEMENT} 
Given a visual observation $O$, the environment description \textit{E} and task instruction \textit{L}, where $O=\{o_1,\dots,o_T\}$ denotes a set of RGB observations. Our goal is to learn a hierarchical dual-arm manipulation policy capable of decomposing the task into structured subtasks, grounding each subtask into executable actions, and allocating these actions to two manipulators in a temporally consistent and collision-free manner. 

The system first predicts a sequence of subtasks $\mathcal{S}_t=\{s_i\}_{i=1}^{N_t} \leftarrow \pi_{\text{subtask}}(o_t, e_t,l_t)$, where each $s_i$ denotes an atomic subtask that cannot be further decomposed (e.g., `pick the red block into the red bowl''). Let $\mathcal{D}$ be the completed set and $\mathcal{S}_t=\{\,i\mid\mathrm{Pre}(i)\subseteq\mathcal{D}\,\}$. For each selected subtask $s_i\in\mathcal{S}_t$, an action policy $\pi_{\text{act}}(o_t,s_i, l_t) \rightarrow a_i=(T^{pk}_i,T^{pl}_i)$ grounds the subtask into pick–place end-effector poses, which serve as the action primitives, consistent with the scene geometry.
We denote the dual-arm allocation policy as 
$\pi_{\text{alloc}} : a_i \rightarrow arm_j,\quad j\in\{1,2\}$,
which assigns each $a_i$ to one of the two manipulators.
% Since the predicted subtask sequence $\mathcal{S}_t=\{s_i\}_{i=1}^{N_t}$ may
% contain both ordered and unordered dependencies, 
Since the predicted subtask sequence $\mathcal{S}_t$ is often linearized without encoding parallelism, it can over-constrain execution and limit bimanual concurrency.
We organize it into a set of
levels $\{L_1,\dots,L_K\}$ such that
\[
\mathcal{S}_t = \bigcup_{k=1}^{K} L_k,\quad
L_k \subseteq \mathcal{S}_t,\quad
L_k \cap L_{k'} = \emptyset\ \text{for } k\neq k',
\]
where the ready set
${L}_k = \{\, i \mid s_i\in\mathcal{S}_t,\ \mathrm{Pre}(i)\subseteq\mathcal{D} \,\}$,
which consists of all indices whose subtasks have all predecessors already
completed. Every subtask $s_i\in L_k$ has all its predecessors in the preceding levels,
i.e.,
$\mathrm{Pre}(i)\subseteq \bigcup_{m<k} L_m$. Subtasks within the same level $L_k$ can be allocated freely to either arm and executed in parallel, whereas subtasks in later levels only enter once all required predecessors have been finished.
% For example, one can have $L_1=\{s_1,s_2,s_3\}$ with no mutual precedence,
% $L_2=\{s_4\}$ with $\mathrm{Pre}(4)=\{1,2,3\}$, and 
% $L_3=\{s_5,s_6,s_7\}$ with $\mathrm{Pre}(5)=\mathrm{Pre}(6)=\mathrm{Pre}(7)=\{4\}$ 
% but no ordering among $s_5,s_6,s_7$. 
The scheduler then seeks the allocation policy that minimizes the motion effort
of both arms while satisfying precedence and collision constraints:
\begin{equation}
\begin{aligned}
\pi_{\text{alloc}}
&=
\arg\min_{\pi_{\text{alloc}}}
\sum_{i\in{L}_t}
\mathrm{Cost}\!\left(\tau(\pi_{\text{alloc}})\right)
\\[3pt]
\text{s.t.}\quad
& \quad \forall s_i\in{L}_t, \tau(\cdot)\in\mathcal{C}_{\mathrm{path}},
\end{aligned}
\end{equation}
where $\tau(\cdot)$ denotes the joint trajectory executed by $arm_j$, and
$\mathcal{C}_{\mathrm{path}}$ is the set of collision-free trajectories.

% We denote the dual-arm allocation policy as $\pi_{\text{alloc}} : a_i \to arm_j, j=\{1,2\},$ which assigns each action to one of the two manipulators. The scheduler then seeks the allocation policy that minimizes the motion effort under collision constraints: 
% \begin{equation} 
% \begin{aligned} 
% \pi_{\text{alloc}}^{*} &= \arg\min_{\pi_{\text{alloc}}} \sum_{i\in\mathcal{I}_t} \mathrm{Cost}\!\left( \tau(\pi_{\text{alloc}}) \right) \\ \text{s.t.} \quad & \tau(\cdot)\in\mathcal{C}_{\mathrm{path}}. 
% \end{aligned}
% \end{equation} 
% Here, $\tau(\cdot)$ denotes the joint trajectory executed by $arm_j$, and $\mathcal{C}_{\mathrm{path}}$ is the set of collision-free trajectories. 

The overall objective is to learn the hierarchical policy \begin{equation} \Pi=\{\pi_{\text{subtask}},\;\pi_{\text{act}},\;\pi_{\text{alloc}}\}, \end{equation} which minimizes the cumulative execution cost across the task while respecting the subtask and exploiting dual-arm parallelism whenever feasible.

\section{THE ExS2D FRAMEWORK}
Here, we introduce \textbf{ExS2D}, a framework that enables explicit subtask decomposition and executable action allocation from human instructions in RGB-D observation. Figure~\ref{fig:1} shows an overview of the framework, highlighting its key modules and processes.

\subsection{Vision-Language Subtask Generation}
\label{sec:4.1}
To learn the $\pi_{\text{subtask}}$, \textbf{ExS2D} employs an Multimodal Large Language Model (MLLM) that takes the visual observation $o_t$, environment description $e_t$, and task instruction $l_t$ as input and produces a structured subtask sequence $\hat{S}_t=\{\hat{s}_1,\dots,\hat{s}_{N_t}\}$.  
This sequence serves as the high-level plan for downstream action grounding and arm allocation.
Effective subtask generation requires the model to build precise visual grounding and understand how actions transform the scene.  
However, generic VLM captioning often omits geometric cues and fine-grained spatial details crucial for manipulation.  
To strengthen multimodal alignment, we train the MLLM using two complementary caption-based supervision signals, enabling its visual representations to integrate seamlessly with the LLM’s reasoning space and supporting reliable task decomposition.

As shown in Figure~\ref{fig:1}a, the first supervision signal focuses on static spatial understanding.  
Given a single observation $o_t$, the model is trained to generate an informative scene description $e_t$ that captures objects, attributes, and spatial relations.  
This encourages the MLLM to internalize scene structure and object layout, providing the perceptual foundation needed for subtask reasoning.  
Formally, the model predicts each token $\hat{e}_{t,k}$ based on previous tokens and visual features $W(f=V(o_t))$, where $V(\cdot)$ is a frozen vision encoder and $W$ is a projection into the language space:
\[
\hat{e}_{t,k}=\mathrm{MLLM}_{tuned}(\{e_{t,1},\dots,e_{t,k-1}\}\mid [W(f=V(o_t))]).
\]
The corresponding perception loss is:
\begin{equation}
\mathcal{L}_{\text{perc}}
=
\sum_{k=1}^{K}
\mathrm{CELoss}(\hat{e}_{t,k}, e_{t,k}).
\end{equation}

The second supervision signal targets dynamic action understanding.  
We provide demonstration sequences $o_{1:T}$ and supervise the model to infer the induced subtask sequence $S_t$, such as identifying which object was moved and how the scene changed.  
This enables the MLLM to capture temporal dependencies and action semantics purely from visual differences.  
That is, given the task-included multimodal input $(o_t,e_t,l_t)$, the predicted subtask tokens $\hat{s}_i$ are generated autoregressively:
\[
\hat{s}_i=
\mathrm{MLLM}_{tuned}(\{s_1,\dots,s_{i-1}\}\mid [o_t,e_t,l_t]).
\]
The task decomposition loss is defined as:
\begin{equation}
\mathcal{L}_{\text{task}}
=
\sum_{n=1}^{N}
\mathrm{CELoss}(\hat{s}_i, s_i).
\end{equation}
% Together, the two training signals allow the MLLM learns to ground visual observations and task instruction, decoding them into subtask sequences.  
% The overall objective is a weighted sum:
The total loss $\mathcal{L}$ combines the two terms by weight factor $\lambda$:
\begin{equation}
\mathcal{L}
=\lambda\mathcal{L}_{\text{perc}}+ \mathcal{L}_{\text{task}}.
\end{equation}
This joint training enables the MLLM to integrate spatial grounding with action-level reasoning, yielding reliable subtask plans for downstream robotic execution.

\begin{table*}
  \centering
  \caption{Performance Comparison of Methods Across Tasks.}
  \resizebox{\textwidth}{!}{
   % 需要 \usepackage{array}
\begin{tabular}{l
 >{\centering\arraybackslash}m{1cm} >{\centering\arraybackslash}m{1cm}
 >{\centering\arraybackslash}m{1cm} >{\centering\arraybackslash}m{1cm}
 >{\centering\arraybackslash}m{1cm} >{\centering\arraybackslash}m{1cm}
 >{\centering\arraybackslash}m{1cm} >{\centering\arraybackslash}m{1cm}
 |>{\centering\arraybackslash}m{1cm} >{\centering\arraybackslash}m{1cm}}

    \toprule
    \multirow{2}[6]{*}{\textbf{Method}} 
    & \multicolumn{2}{c}{\makecell{\textbf{Stacking of} \\ \textbf{Blocks in Zone (a)}}} 
    & \multicolumn{2}{c}{\makecell{\textbf{Placement of} \\ \textbf{Blocks in Bowls (b)}}} 
    & \multicolumn{2}{c}{\makecell{\textbf{Finding Horizontal}\\\textbf{Symmetry Letters (c)}}}
    & \multicolumn{2}{c}{\makecell{\textbf{Assembling} \\ \textbf{of Kits (d)}}}
    & \multicolumn{2}{c}{\textbf{Macro-Average}}  \\
    \cmidrule(lr){2-11}          
          & \textbf{SR$\uparrow$}  & \textbf{Step$\downarrow$}  
          & \textbf{SR$\uparrow$}  & \textbf{Step$\downarrow$}  
          & \textbf{SR$\uparrow$}  & \textbf{Step$\downarrow$}  
          & \textbf{SR$\uparrow$}  & \textbf{Step$\downarrow$}  
          & \textbf{SR$\uparrow$}  & \textbf{Step$\downarrow$}  \\
    \midrule 
                        
    \textbf{CLIPort}    & \textbf{87.82}& 7.25         
                        & 91.25         & 7.23          
                        & \textbf{80.01}& 7.01          
                        & \textbf{95.21}& 7.80          
                        & 88.57         & 6.27  \\         
                        
    \midrule                    
    \textbf{LLM+MAP}    & 43.49         & 4.12           
                        & 87.58         & 3.21          
                        & 71.25         & 3.82            
                        & 78.55         & 2.58           
                        & 70.22         & 3.43  \\
                        
    \textbf{RoCo}       & 48.02         & 4.01           
                        & 88.57         & 3.25        
                        & 73.74         & 3.41      
                        & 84.78         & 2.48         
                        & 73.78         & 3.29  \\ 
    \midrule                    
    \textbf{ExS2D}      & 85.28         & \textbf{3.11}     
                        & \textbf{92.01}& \textbf{3.12}  
                        & 78.89         & \textbf{2.87} 
                        & 93.25         & \textbf{2.35}     
                        & 87.36         & \textbf{2.86}   \\
    \bottomrule
    \end{tabular}%
 }
\label{tab:tab1}%
\vspace{-1mm}
\end{table*}

\subsection{Subtask Guided Action Mapping}
\label{sec:4.2}
The \textbf{ExS2D} implements the action policy
$\pi_{\text{act}}$, which grounds each predicted subtask $s_i$ into an
executable action $a_i = (T^{pk}_i, T^{pl}_i)$ consistent with the
scene geometry.  
To achieve this grounding, the module combines a subtask-conditioned vision–language signal with CLIPort.
Specifically, given a cropped observation and a textual query, a VLM predicts an object-level bounding box, which is further refined into a segmentation mask via SAM.
This mask is then used to guide CLIPort’s attention, producing spatially focused probability maps that highlight task-relevant regions for pick and place, as illustrated in Figure~\ref{fig:1}b.
To train the CLIPort-based action generator $\pi_{\text{act}}$, we use oracle actions as supervision.  
Each training sample provides the visual observation $o_t$, the grounded
subtask $s_i$, and the task instruction $l_t$.  
% The model predicts an action $\hat{a}_t = p_{\psi}(a_t \mid o_t, s_i, l_t)$,
% which is supervised with an $\ell_2$ regression loss:
The model predicts an action $\hat{a}_t$, which is supervised with an $\ell_2$ loss:
\begin{equation}
\mathcal{L}_{\text{act}}
= \sum_t \lVert a_t - \hat{a}_t \rVert_2.
\end{equation}

After training, CLIPort computes the $i_{th}$ pick–place pair action $a_i$ from the  $o_t$,  $l_t$ and  $s_i$:
\begin{equation}\label{eq:5}
T^{pk}_i = \arg\max_{(u,v)}
Q_{\text{pick}}((u,v)\mid o_t, l_t, s_i),
\end{equation}
\begin{equation}\label{eq:6}
T^{pl}_i = \arg\max_{\Delta\tau}
Q_{\text{place}}(\Delta\tau \mid o_t, T^{pk}_i, l_t, s_i),
\end{equation}
where $Q_{\text{pick}}$ and $Q_{\text{place}}$ are CLIPort’s learned spatial
Q-maps, $(u,v)$ corresponds to a 3D location in the orthographic heightmap
$o_t$, and $\Delta\tau \in SE(2)$ denotes a candidate placement transformation.
To refine grounding for additional objects referenced in the subtask, we apply a
mask-guided visual refinement step.  
A localized crop $O_c$ around $(T^{pk}_i, T^{pl}_i)$ is queried with VLM to obtain
a bounding box, which is further processed by SAM to produce a fine-grained
object mask.  
Applying this mask yields a refined observation $o_m$, which selectively
preserves the geometry of the queried object while suppressing irrelevant scene
content.
CLIPort is then reapplied on the masked observation to infer the next pick–place
action:
\begin{equation}\label{eq:8}
T^{pk}_{i+1} = \arg\max_{(u,v)}
Q_{\text{pick}}((u,v)\mid o_m, l_t, s_{i+1}),
\end{equation}
\begin{equation}\label{eq:9}
T^{pl}_{i+1} = \arg\max_{\Delta\tau}
Q_{\text{place}}(\Delta\tau \mid o_m, T^{pk}_{i+1}, l_t, s_{i+1}).
\end{equation}

This mask-guided refinement couples semantic grounding with the spatial precision of CLIPort, enabling $\pi_{\text{act}}$ to
robustly localize multiple objects and produce actions for ubtasks.

\subsection{Precedence-aware Dual-Arm Coordinator }
\label{sec:4.3}
To realize the allocation policy $\pi_{\text{alloc}}$, we employ an MLLM-driven allocator, \textbf{P-DCoord}, that reasons over precedence constraints in the predicted $\mathcal{S}_t$ and partitions it into a sequence of $\{L_k\}$,
with $L_k$ containing all subtasks whose predecessors have been completed
as shown in Figure~\ref{fig:1}c.
This precedence inference is conditioned on the $e$ and $o$, enabling the allocator to reason about subtask executability under the current environment.
Within a given ${L}_t$, subtasks have no temporal
constraints  and can therefore be executed in parallel at the current step.  
Each subtask $s_i\in{L}_t$ is grounded into a concrete action $a_i$ by the\textbf{ SA-Map}.

Since the number of subtasks within a single ready set is typically small,
we adopt a greedy-based allocation strategy to assign the resulting actions
to the two manipulators.  
For a sequence of grounded action $a_i$, the coordinator enumerates
candidate arm assignments and estimates their execution cost based on the
current end-effector poses $e_k$ of arm $k$.  
The heuristic cost consists of a movement cost
% $c_{mv}(i,k)=\|e_k - T^{pk}_i\|_2$
$\|e_k - T^{pk}_i\|_2$,
which measures the effector to reach the pick pose, and a transfer cost
% $c_{tf}(i)=\|T^{pk}_i - T^{pl}_i\|_2$
$\|T^{pk}_i - T^{pl}_i\|_2$,
which captures the cost from pick to place.  
These terms together provide a approximation of the execution cost
for allocating action $a_i$ to arm $k$.
Based on the heuristic cost estimation, \textbf{P-DCoord} narrows the allocation space to a small set of promising dual-arm assignments.  
For each candidate $\pi_{\text{alloc}}$, a synchronized RRT* planner generates
dual-arm trajectories $\tau(\pi_{\text{alloc}})$, which are discretized and
checked for arm–arm and arm–environment collisions.  
If all candidates are infeasible, the system falls back to asynchronous
execution and replans the allocation after resetting the arm.
 
\begin{figure}[h]
  \centering
  \includegraphics[width=1\linewidth]{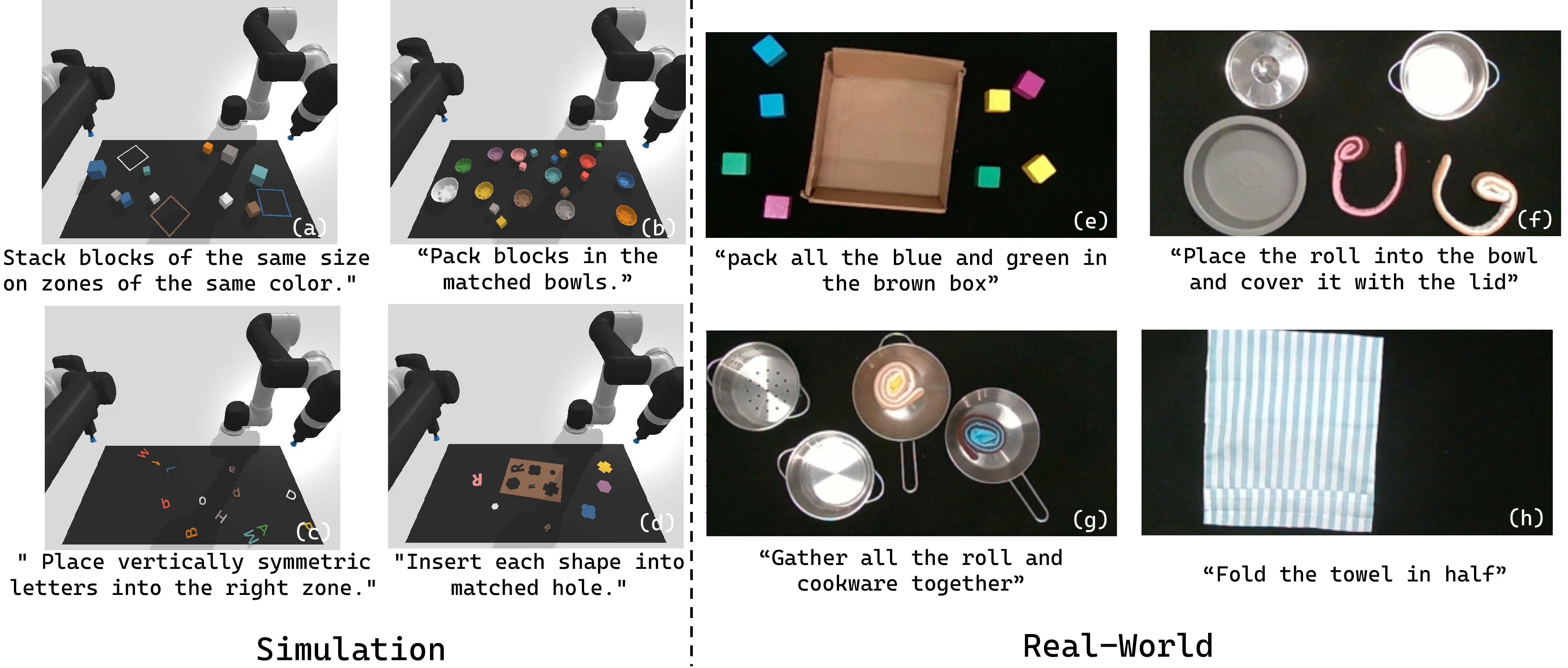}
  \caption{Language-Conditioned Manipulation Tasks and Setup. We evaluated on four dual-arm simulation tasks (a–d)  and four real-world tasks (e–h) on the Elephant manipulator.}
  % We evaluated on four dual-arm simulation tasks (a–d) in Ravens~\cite{zeng2021transporter} and four real-world tasks (e–h) on the Elephant manipulator.}
  \label{fig:2}
\vspace{-5mm}
\end{figure}
%%%%%%%%%%%%%%%%%%%%%%%%%%%%%%%%%%%%%%%%%%%%%%%%%%%%%%%%%%%%%%%%%%%%%%%%
\section{EXPERIMENTS}
\subsection{Experiments Setup}

\noindent\textbf{Task and Environment.} We evaluate ExS2D on four language-conditioned simulation tasks from the Raven Bench ~\cite{pmlr-v164-shridhar22a} and four real-world dual-arm tasks, as shwon in Fig~\ref{fig:2}. Real-world experiments are conducted on two Elephant Pro 630 robots with parallel grippers, using a top-down RealSense D455 camera with downsampled 640$\times$480 images.

\noindent\textbf{Implementation detail.} \textbf{VL-SubGen} fine-tunes Qwen2.5-VL-7B with LoRA on the visual and cross-modal projection layers as well as the attention components. 
In \textbf{SA-Map}, CLIPort~\cite{pmlr-v164-shridhar22a} is trained for action mapping, while OWL-ViT~\cite{Heigold_2023_ICCV}  is adopted as the VLM and a lightweight SAM~\cite{kirillov2023segment} is used during action generation to produce masks for refined target selection.  
For the \textbf{P-DCoord} phase, we employ a pretrained Qwen3-VL as an external reasoner to perform online inference over task precedence and candidate action pairs.

\noindent\textbf{Evaluation Metrics and Baselines.}
We report \textbf{Success rate (SR)} (0-100, with partial credit for incomplete completion) and \textbf{Step}, the number of dual arms execution iterations; for failed executions, steps are computed using single-arm plans for fair comparison.
We additionally report \textbf{Token Accuracy} to evaluate the correctness of the generated subtask sequences, and \textbf{Cost} defined as the sum of Euclidean distances between each action target and the current end-effector position.
For baselines, we group methods by planning strategy, End-to-End: \textbf{CLIPort}~\cite{pmlr-v164-shridhar22a}, and MLLM-based Planners: \textbf{PAR}~\cite{zhang2023lohoravenslonghorizonlanguageconditionedbenchmark}, \textbf{LLM+MAP}~\cite{chu2025llmmapbimanualrobottask}, and \textbf{RoCo}~\cite{10610855}.

\begin{figure} 
\centering
\includegraphics[width=0.9\linewidth]{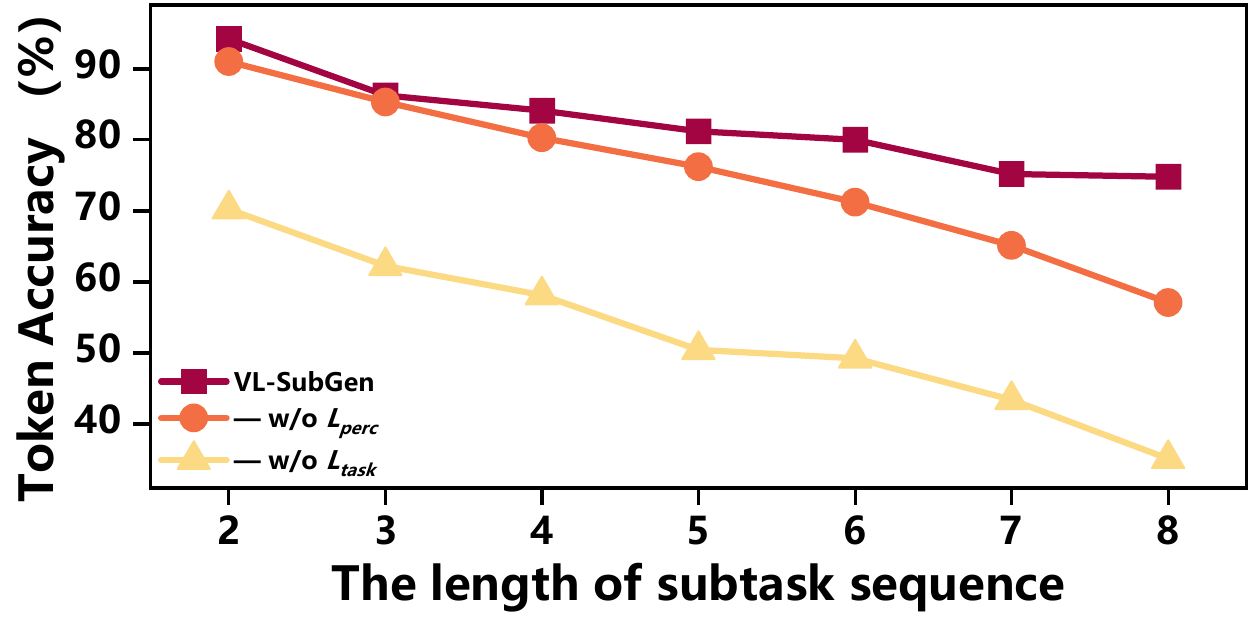} % Adjusted width
\caption{Token accuracy versus subtask sequence length.}
\label{fig:3}
\end{figure}

\begin{table}
  \centering
  \caption{Ablation Studies}
  \begin{tabular}{l l cc}
    \toprule
    \textbf{Component} &   & \textbf{SR} $\uparrow$ & \textbf{Cost} $\downarrow$ \\
    \midrule
    % \multirow{1}{*}{VL-SubGen} 
    %   & - w/o VL-SubGen              & 30.21 & -- \\
    % \midrule
    \multirow{2}{*}{SA-Map} 
      & - w/o Mask                   & 70.21 & -- \\
      & - w Fixed Mask              & 78.21 & -- \\
    \midrule

    \multirow{2}{*}{P-DCoord} 
      & - w/o Precedence Reasoning   & 63.09 & 2.24 \\
      & - w Fixed Allocation        & 80.21 & 3.58 \\
    \midrule 
     \multirow{2}{*}{\textbf{Full Model}} 
      & \textbf{Our Method}        & 87.36 & \textbf{2.02} \\
      % & - w subtask oracle         & \textbf{90.25} & - \\
      % & - w action oracle          &   84.84    & -\\
       & - w  oracle                & \textbf{92.57}   & -\\
    \bottomrule
  \end{tabular}
  \label{tab:tab2}
  \vspace{-5mm}
\end{table}

\begin{figure}[t]
    \centering
    % --- (a) Token Accuracy ---
    \begin{subfigure}[t]{0.48\linewidth}
        \centering
        \includegraphics[width=\linewidth]{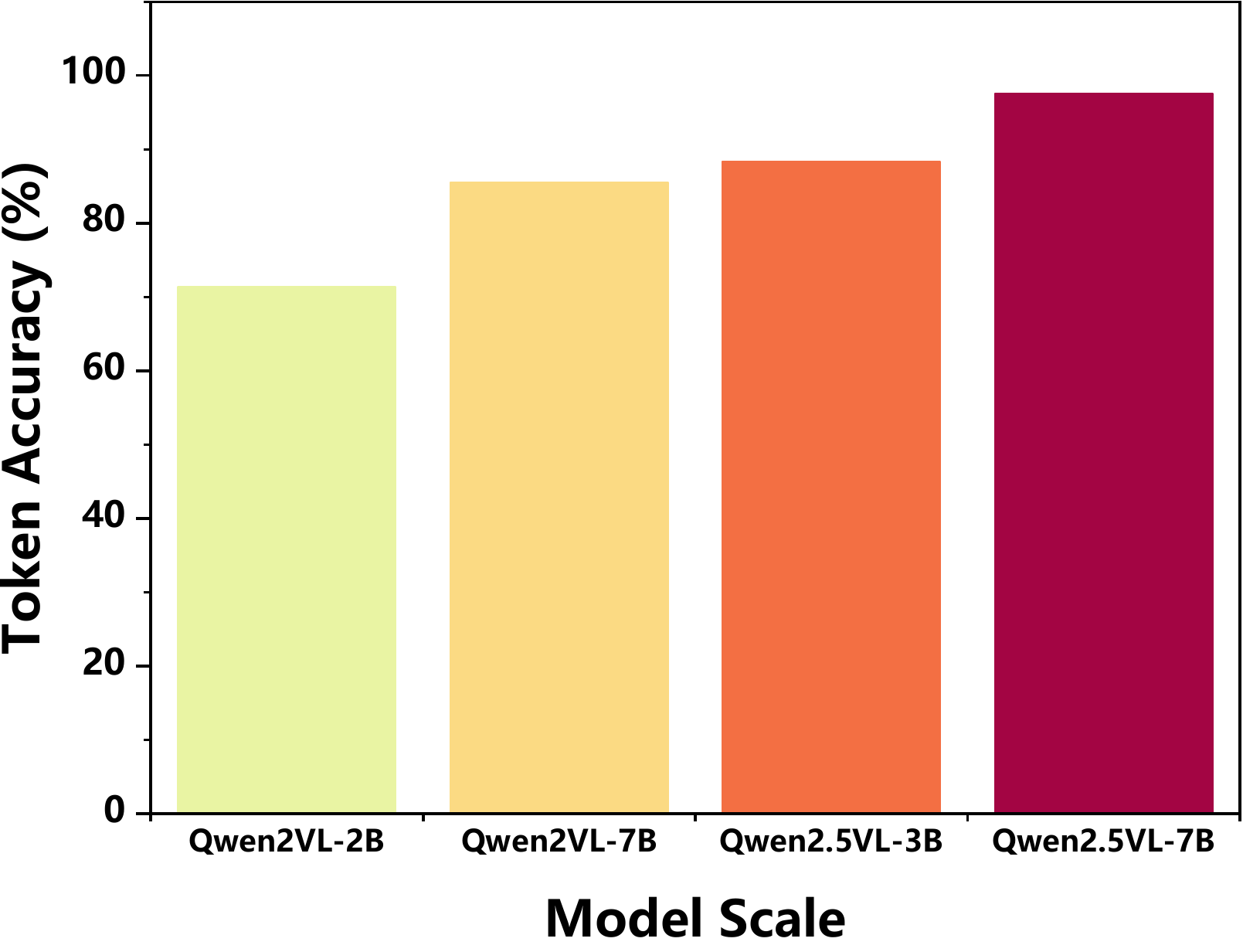}
        \caption{}
        \label{fig:token-acc}
    \end{subfigure}
    \hfill
    % --- (b) Time Cost Breakdown ---
    \begin{subfigure}[t]{0.48\linewidth}
        \centering
        \includegraphics[width=\linewidth]{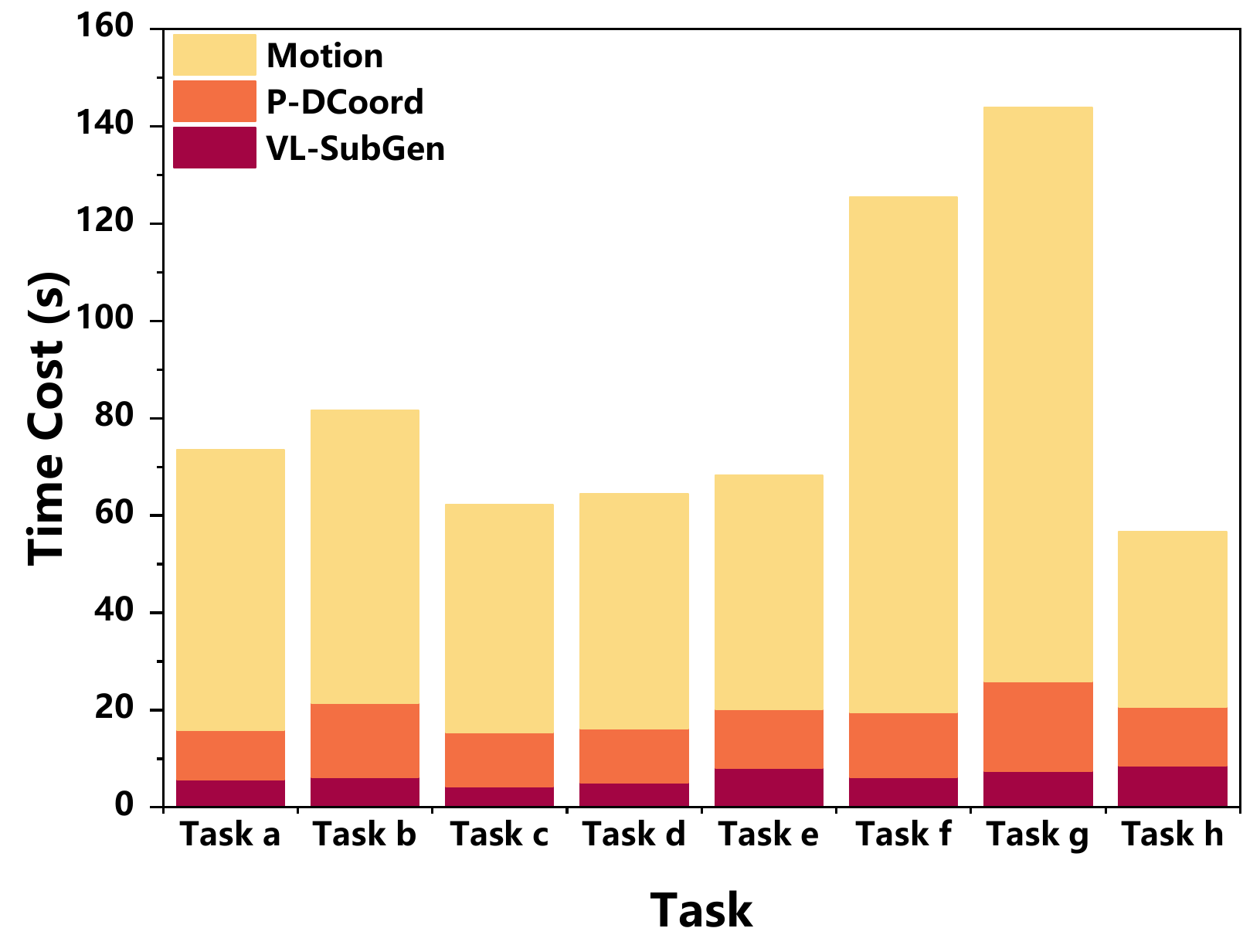}
        \caption{}
        \label{fig:time-cost}
    \end{subfigure}

    \caption{
    % Model capability and efficiency analysis. 
    (a) Token accuracy for different Model scales. 
    (b) Time cost inference over tasks.}
    \label{fig:acc-and-time}
\end{figure}

\subsection{Quantitative Results}
Table~\ref{tab:tab1} summarizes results on four tasks. 
Compared to the single-arm baseline CLIPort, \textbf{ExS2D} reduces the average steps by 54.4\% with only a 1.22\% drop in success rate, demonstrating an efficiency--accuracy trade-off. 
Against dual-arm planners, \textbf{ExS2D} attains higher success with fewer steps. 
On task (a) with strong temporal dependencies, precedence-aware allocation yields a clear gain over LLM+MAP and RoCo. 
% Overall, \textbf{ExS2D} achieves a better balance between efficiency and success by jointly generating actions and reasoning over inter-subtask dependencies.
Overall, \textbf{ExS2D} balances efficiency and success through joint action generation and subtask dependency reasoning.

\begin{table}[t]
  \centering
  \caption{Real-world task results.}
  \begin{threeparttable}
  \begin{tabular}{cccc}
    \toprule
    \textbf{Task} & \makecell{\textbf{Single-arm Samples}\\\textbf{(Train/Test)}} & \textbf{Bimanual Samples} & \textbf{SR} $\uparrow$\\
    \midrule
    e & $30\,(97)\ /\ 5\,(21)$   & 0 & $70.33$\\
    f & $50\,(165)\ /\ 5\,(18)$  & 0 & $66.67$\\
    g & $40\,(114)\ /\ 6\,(22)$  & 0 & $42.24$\\
    h & $35\,(102)\ /\ 8\,(24)$  & 0 & $61.78$\\
    \bottomrule
  \end{tabular}
  % \begin{tablenotes}[flushleft]\footnotesize
  %   \item \textit{Train/Test are reported as single-arm samples (image--action pairs in parentheses). No bimanual demonstrations are used.}
  % \end{tablenotes}
  \end{threeparttable}
  \label{tab:tab3}
  \vspace{-5mm}
\end{table}

\begin{figure}[h]
\centering
\includegraphics[width=1\linewidth]{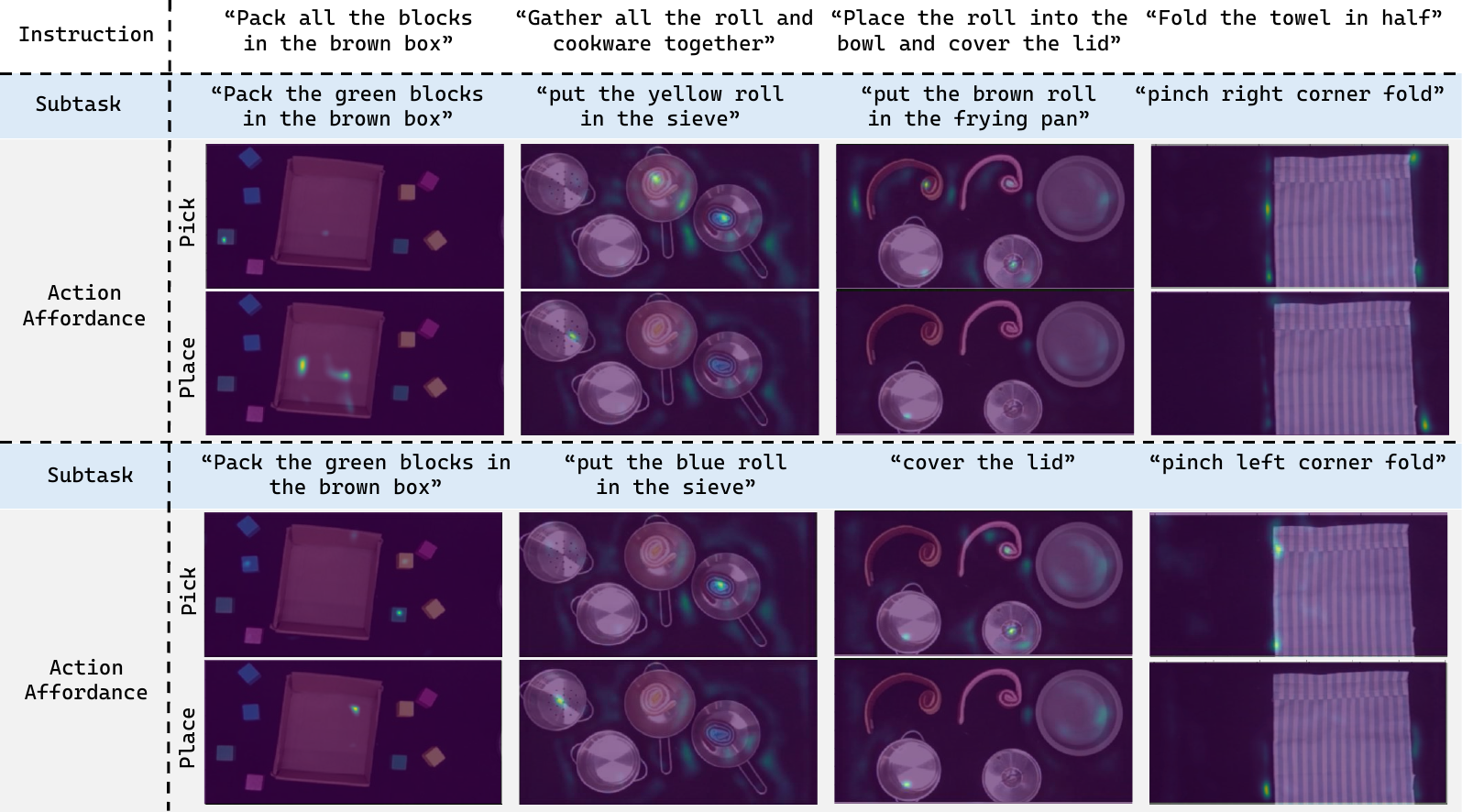} % Adjusted width
\caption{The affordance predictions for each real-world task.}
\label{fig:4}
\vspace{-5mm}
\end{figure}

% \subsection{Ablation Study}
% We conduct ablations on the action generation and allocation components in Table~\ref{tab:tab2}. 
% Removing the attention extension collapses the macro-average SR to 49.13\%, while extending attention on the global observation recovers performance to 68.08\%. 
% Replacing semantic localization with a fixed 64$\times$64 crop further improves SR to 72.17\%, and combining semantic cropping with mask-guided exclusion achieves the highest SR of 73.43\%. 
% On the allocation side, disabling collision awareness reduces SR to 63.09\% and increases allocation time and arm motion, while kinematic simplification and nearest-neighbor heuristics further degrade efficiency. 
% Overall, LDAC’s integrated cost model yields the highest success rate with the fastest allocation and shortest trajectories.

\subsection{Ablation Study}
We conduct ablation experiments to quantify the contribution of each module in \textbf{ExS2D}. 
% We analyze the reasoning robustness of \textbf{VL-SubGen} in Figure~\ref{fig:3}, and report success rate and execution cost in Table~\ref{tab:tab2}.

\noindent\textbf{Reasoning robustness of VL-SubGen.}
As the subtask sequence length increases in Figure~\ref{fig:3}, token accuracy consistently decreases for all variants. Importantly, the full \textbf{VL-SubGen} remains the most robust across lengths, while removing $L_{\text{perc}}$ leads to a larger degradation, and removing $L_{\text{task}}$ results in the steepest drop. This verifies that both $L_{\text{perc}}$ and $L_{\text{task}}$ are necessary to maintain stable reasoning for task.

\noindent\textbf{Effect of SA-Map masking.}
As shown in Table~\ref{tab:tab2}, removing the mask guidance (w/o Mask) causes a clear $17.15\%$ SR decrease, indicating that the action grounding becomes less reliable without suppressing irrelevant regions. Using a fixed mask (w/ Fixed Mask) partially recovers the performance to $78.21\%$ SR, underscoring
the importance of adaptive semantic masking for robust grounding.

\noindent\textbf{Effect of P-DCoord allocation.}
Without Precedence Reasoning reduces SR to $63.09\%$ and raises cost to $2.24$, confirming the importance of dependency-aware coordination. Fixed  arm order (w Fixed Allocation) improves SR to $80.21\%$ but increases 1.56 cost, implying that static assignment can complete tasks but sacrifices efficiency due to reduced parallelism. 

Overall, our method achieves the best trade-off, and the oracle setting further boosts SR to $92.57\%$, indicating additional headroom mainly comes from improving upstream predictions.
% and grounding quality.

\subsection{Further Analysis}
Figure~\ref{fig:acc-and-time} reports both model capability and inference efficiency.
In Figure~\ref{fig:acc-and-time}(a), token accuracy increases with model scale; Qwen2.5VL consistently outperforms Qwen2VL, with Qwen2.5VL-7B performing best. 
In Figure~\ref{fig:acc-and-time}(b), inference time is decomposed into Motion, VL-SubGen, and P-DCoord; Motion dominates while the other two add minor but non-negligible overhead, suggesting improved performance.
% with limited efficiency loss.

\subsection{Real-Robot Experiments}
Table~\ref{tab:tab3} summarizes the real-robot results on four tasks e--h. \textbf{ExS2D} achieves competitive success rates with $60.51\%$ under few-shot training. Notably, tasks g and h involve temporal dependencies among subtasks (e.g., ``place the roll into the bowl'' must precede ``cover the lid'', and towel folding requires corner folds simultaneously). Moreover, task h requires bimanual coordination during folding, yet our framework accomplishes it using only few-shot \textit{single-arm} demonstrations (image--action pairs), without any bimanual demonstrations or dual-arm datasets. Fig.~\ref{fig:4} visualizes representative pick/place affordance predictions conditioned on subtasks, illustrating that \textbf{ExS2D} localizes task-relevant regions and generates executable actions even with limited data.

%%%%%%%%%%%%%%%%%%%%%%%%%%%%%%%%%%%%%%%%%%%%%%%%%%%%%%%%%%%%%%%%%%%%%%%%
\section{Conclusion and Limitations}
We present \textbf{ExS2D}, a hierarchical dual-arm manipulation framework that decomposes a task into structured subtasks, grounds each subtask into executable pick--place actions, and allocates actions to two arms under precedence and collision constraints.
While \textbf{VL-SubGen} produces structured subtasks and \textbf{SA-Map} grounds them into executable actions, the reasoning over temporal dependencies and execution order is handled by \textbf{P-DCoord}, which evaluates candidate action pairs under explicit precedence and motion-planning constraints.
Extensive simulation and real-robot results demonstrate that \textbf{ExS2D} improves execution efficiency while maintaining competitive success, and ablations verify the importance of semantic masking, and precedence reasoning for robustness.

Despite these gains, the current system adopts an open-loop execution paradigm based on end-effector abstractions and mainly supports planar pick–place actions, without closed-loop feedback for fine-grained bimanual interaction. Consequently, improvements in dexterous manipulation remain limited. Future work will explore closed-loop control with richer dexterous skills and extend to multi-arm scenarios.
\section*{Acknowledgments}
This work was supported by the National Natural Science Foundation of China under Grants 62322601 and 62572084, and the Fundamental Research Funds for the Central Universities (No.2024IAIS-QN017, 2025CDJZDGF001).
\bibliographystyle{IEEEbib}
\bibliography{icme2026references}

@InProceedings{Heigold_2023_ICCV,
    author    = {Heigold, Georg and Minderer, Matthias and Gritsenko and others},
    title     = {{V}ideo {OWL-V}i{T}: {T}emporally-consistent {O}pen-world {L}ocalization in {V}ideo},
    booktitle = {Proceedings of the IEEE/CVF International Conference on Computer Vision (ICCV)},
    month     = {October},
    year      = {2023},
    pages     = {13802-13811}
}

@ARTICLE{11075556,
  author={Tu, Yuyang and Wang, Yunlong and Zhang, Hui and others},
  journal={IEEE Robotics and Automation Letters (RAL)}, 
  title={{L}anguage-{E}mbedded {6D} {P}ose {E}stimation for {T}ool {M}anipulation}, 
  year={2025},
  volume={10},
  number={9},
  pages={8618-8625},
  doi={10.1109/LRA.2025.3587559}}

@ARTICLE{10900471,
  author={Wen, Junjie and Zhu and others},
  journal={IEEE Robotics and Automation Letters (RAL)}, 
  title={{TinyVLA}: {T}oward {F}ast, {D}ata-{E}fficient {V}ision-{L}anguage-{A}ction {M}odels for {R}obotic {M}anipulation}, 
  year={2025},
  volume={10},
  number={4},
  pages={3988-3995},
  doi={10.1109/LRA.2025.3544909}}

@article{bai2024twostep,
  title={{T}wostep: {M}ulti-agent {T}ask {P}lanning using {C}lassical {P}lanners and {L}arge {L}anguage {M}odels},
  author={Bai, David and Singh, Ishika and Traum, David and Thomason, Jesse},
  journal={arXiv preprint arXiv:2403.17246},
  year={2024}
}

@INPROCEEDINGS{10610855,
  author={Mandi and others},
  booktitle={2024 IEEE International Conference on Robotics and Automation (ICRA)}, 
  title={{R}o{C}o: {D}ialectic {M}ulti-{R}obot {C}ollaboration with {L}arge {L}anguage {M}odels}, 
  year={2024},
  volume={},
  number={},
  pages={286-299},
 
  doi={10.1109/ICRA57147.2024.10610855}}

@article{chu2025llmmapbimanualrobottask,
  title={{LLM+MAP}: {B}imanual {R}obot {T}ask {P}lanning using {L}arge {L}anguage {M}odels and {P}lanning {D}omain {D}efinition {L}anguage},
  author={Kun Chu and Xufeng Zhao and Cornelius Weber and Stefan Wermter},
  journal={ArXiv},
  year={2025},
  volume={abs/2503.17309},
  url={https://api.semanticscholar.org/CorpusID:277244500}
}

@ARTICLE{10323148,
  author={Purushottam and others},
  journal={IEEE Robotics and Automation Letters (RAL)}, 
  title={{D}ynamic {M}obile {M}anipulation via {W}hole-Body {B}ilateral {T}eleoperation of {a} {W}heeled {H}umanoid}, 
  year={2024},
  volume={9},
  number={2},
  pages={1214-1221},
}

@InProceedings{pmlr-v205-shaw23a,
  title = 	 {{V}ideo{D}ex: {L}earning {D}exterity from {I}nternet {V}ideos},
  author =       {Shaw, Kenneth and Bahl, Shikhar and Pathak, Deepak},
  booktitle = 	 {Proceedings of The 6th Conference on Robot Learning},
  pages = 	 {654--665},
  year = 	 {2023},
  volume = 	 {205},
  series = 	 {Proceedings of Machine Learning Research (PMLR)},
  month = 	 {14--18 Dec},
  publisher =    {PMLR}
}

@InProceedings{pmlr-v164-jang22a,
  title = 	 {{BC-Z}: {Z}ero-{S}hot {T}ask {G}eneralization with {R}obotic {I}mitation {L}earning},
  author =       {Jang, Eric and Irpan and others},
  booktitle = 	 {Proceedings of the 5th Conference on Robot Learning},
  pages = 	 {991--1002},
  year = 	 {2022},
  volume = 	 {164},
  series = 	 {Proceedings of Machine Learning Research (PMLR)},
  month = 	 {08--11 Nov},
  publisher =    {PMLR}
}

@inproceedings{chi2023diffusion,
	title={{D}iffusion {P}olicy: {V}isuomotor {P}olicy {L}earning via {A}ction {D}iffusion},
	author={Chi, Cheng and Feng, Siyuan and Du, Yilun and Xu, Zhenjia and Cousineau, Eric and Burchfiel, Benjamin and Song, Shuran},
	booktitle={Proceedings of Robotics: Science and Systems (RSS)},
	year={2023}
}

@INPROCEEDINGS{wu2024gellogenerallowcostintuitive,
  author={Wu, Philipp and Shentu, Yide and Yi, Zhongke and others},
  booktitle={2024 IEEE/RSJ International Conference on Intelligent Robots and Systems (IROS)}, 
  title={{GELLO}: {A} {G}eneral, {L}ow-{C}ost, and {I}ntuitive {T}eleoperation {F}ramework for {R}obot {M}anipulators}, 
  year={2024},
  volume={},
  number={},
  pages={12156-12163},
  doi={10.1109/IROS58592.2024.10801581}}

@inproceedings{kirillov2023segment,
  title={{S}egment {A}nything},
  author={Kirillov, Alexander and Mintun, Eric and Ravi, Nikhila and others},
  booktitle={Proceedings of the IEEE/CVF international conference on computer vision (ICCV)},
  pages={4015--4026},
  year={2023}
}

@inproceedings{fu2024mobilealohalearningbimanual,
  author    = {Fu, Zipeng and Zhao, Tony Z. and Finn, Chelsea},
  title={{M}obile {ALOHA}: {L}earning {B}imanual {M}obile {M}anipulation with {L}ow-{C}ost {W}hole-{B}ody {T}eleoperation}, 
  booktitle = {{Conference on Robot Learning (CoRL)}},
  year      = {2024},
}

@INPROCEEDINGS{zhao2023learningfinegrainedbimanualmanipulation, 
    AUTHOR    = {Tony Z. Zhao AND Vikash Kumar AND Sergey Levine AND Chelsea Finn}, 
    TITLE     = {{L}earning {F}ine-{G}rained {B}imanual {M}anipulation with {L}ow-{Co}st {H}ardware}, 
    BOOKTITLE = {Proceedings of Robotics: Science and Systems (RSS)}, 
    YEAR      = {2023}, 
    DOI       = {10.15607/RSS.2023.XIX.016} 
}

@inproceedings{joublin2023copalcorrectiveplanningrobot,
   title={{CoPAL}: {C}orrective {P}lanning of {R}obot {A}ctions with {L}arge {L}anguage {M}odels},
   url={http://dx.doi.org/10.1109/ICRA57147.2024.10610434},
   DOI={10.1109/icra57147.2024.10610434},
   booktitle={2024 IEEE International Conference on Robotics and Automation, ICRA},
   publisher={IEEE},
   author = {Joublin, Frank and Ceravola, Antonello and Smirnov, Pavel and others},
   year={2024},
   month=may, pages={8664–-8670} }

@article{liu2023llmpempoweringlargelanguage,
  title={{LLM+P}: {E}mpowering {L}arge {L}anguage {M}odels with {O}ptimal {P}lanning Proficiency},
  author={Liu, Bo and Jiang, Yuqian and Zhang, Xiaohan and others},
  journal={arXiv preprint arXiv:2304.11477},
  year={2023}
}

@InProceedings{Wang_2024_CVPR,
    author    = {Wang, Jun and others},
    title     = {{C}yber{D}emo: {A}ugmenting {S}imulated {H}uman {D}emonstration for {R}eal-{W}orld {D}exterous {M}anipulation},
    booktitle = {Proceedings of the IEEE/CVF Conference on Computer Vision and Pattern Recognition (CVPR)},
    month     = {June},
    year      = {2024},
    pages     = {17952-17963}
}

@ARTICLE{10685120,
  author={Zhou, Hongkuan and others},
  journal={IEEE Robotics and Automation Letters (RAL)}, 
  title={{L}anguage-{C}onditioned {I}mitation {L}earning {W}ith {B}ase {S}kill {P}riors {U}nder {U}nstructured {D}ata}, 
  year={2024},
  volume={9},
  number={11},
  pages={9805-9812},
}

@article{zhang2023lohoravenslonghorizonlanguageconditionedbenchmark,
   title={{L}o{H}o{R}avens: {A} {L}ong-{H}orizon {L}anguage-{C}onditioned {B}enchmark for {R}obotic {T}abletop {M}anipulation}, 
  author={Shengqiang Zhang and Philipp Wicke and L{\"u}tfi Kerem Senel and Luis F. C. Figueredo and Abdeldjallil Naceri and Sami Haddadin and Barbara Plank and Hinrich Sch{\"u}tze},
  journal={ArXiv},
  year={2023},
  volume={abs/2310.12020},
  url={https://api.semanticscholar.org/CorpusID:264289288}
}

@article{wu2024fastumiscalablehardwareindependentuniversal,
  title={{Fast-UMI}: {A} {S}calable and {H}ardware-{I}ndependent {U}niversal {M}anipulation {I}nterface},
  author={Wu, Ziniu and Wang, Tianyu and Guan, Chuyue and Jia, Zhongjie and others},
  journal={arXiv e-prints},
  pages={arXiv--2409},
  year={2024}
}

@article{JUNGBLUTH2022156,
title = {Reinforcement Learning-based Scheduling of a Job-Shop Process with Distributedly Controlled Robotic Manipulators for Transport Operations},
journal = {IFAC-PapersOnLine},
volume = {55},
number = {2},
pages = {156-162},
year = {2022},
author = {Simon and others},
}

@inproceedings{jiang2023vimageneralrobotmanipulation,
  author       = {Yunfan Jiang and
                  Agrim Gupta and
                  Zichen Zhang and others},
  title        = {{VIMA:} {R}obot {M}anipulation with {M}ultimodal {P}rompts},
  booktitle    = {International Conference on Machine Learning, {ICML}},
  volume       = {202},
  pages        = {14975--15022},
  year         = {2023},
}

@INPROCEEDINGS{zhang2024learningdualarmobjectrearrangement,
  author={Zhang, Shishun and She, Qijin and others},
  booktitle={2024 IEEE International Conference on Robotics and Automation (ICRA)}, 
  title={{L}earning {D}ual-arm {O}bject {R}earrangement for {C}artesian {R}obots}, 
  year={2024},
  volume={},
  number={},
  pages={7440-7446},
  doi={10.1109/ICRA57147.2024.10610573}}

@INPROCEEDINGS{li2024synchronizeddualarmrearrangementcooperative,
  author={Li, Wenhao and Zhang, Shishun and Dai, Sisi and Huang, Hui and Hu, Ruizhen and Chen, Xiaohong and Xu, Kai},
  booktitle={2024 IEEE International Conference on Robotics and Automation (ICRA)}, 
  title={{S}ynchronized {D}ual-arm {R}earrangement via {C}ooperative m{TSP}}, 
  year={2024},
  volume={},
  number={},
  pages={9242-9248},
  doi={10.1109/ICRA57147.2024.10610424}}

@InProceedings{pmlr-v164-shridhar22a,
  title = 	 {{CLIP}ort: {W}hat and {W}here {P}athways for {R}obotic {M}anipulation},
  author =       {Shridhar, Mohit and Manuelli, Lucas and others},
  booktitle = 	 {Proceedings of the 5th Conference on Robot Learning, CoRL},
  pages = 	 {894--906},
  year = 	 {2022},
publisher =    {PMLR},
  volume = 	 {164},
}

@article{yang2024bestmanmodularmobilemanipulator,
  author  = {Kui Yang and Nieqing Cao and others},
  title   = {{B}est{M}an: {A} {M}odular {M}obile {M}anipulator {P}latform for {E}mbodied {AI} with {U}nified {S}imulation-hardware {API}s},
  journal = {Frontiers of Computer Science},
  year    = {2025},
  volume  = {19},
  number  = {9},
  pages   = {199361},
  doi     = {10.1007/s11704-025-41109-6}
}

@article{liu2026fly,
  title={{O}n-the-{F}ly {VLA} {A}daptation via {T}est-{T}ime {R}einforcement {L}earning},
  author={Liu, Changyu and Liu, Yiyang and Wang, Taowen and others},
  journal={arXiv preprint arXiv:2601.06748},
  year={2026}
}

@inproceedings{wang2025exploring,
  title={{E}xploring the {A}dversarial {V}ulnerabilities of {V}ision-{L}anguage-{A}ction {M}odels in {R}obotics},
  author={Wang, Taowen and Han, Cheng and Liang, James and Yang, Wenhao and Liu, Dongfang and Zhang, Luna Xinyu and Wang, Qifan and Luo, Jiebo and Tang, Ruixiang},
  booktitle={Proceedings of the IEEE/CVF International Conference on Computer Vision (ICCV)},
  pages={6948--6958},
  year={2025}
}

@inproceedings{su2025ova,
  title={{OVA-F}ields: {W}eakly {S}upervised {O}pen-Vocabulary {A}ffordance {F}ields for {R}obot {O}perational {P}art {D}etection},
  author={Su, Heng and Xie, Mengying and Cao, Nieqing and others},
  booktitle={Proceedings of the IEEE/CVF International Conference on Computer Vision, ICCV},
  pages={6385--6395},
  year={2025}
}

\end{document}